\documentclass[11pt]{article}
\usepackage{coling2020}
\usepackage{times}
\usepackage{url}
\usepackage{pgf-pie}
\usepackage{caption}
\usepackage{latexsym}

\title{Inno at SemEval-2020 Task 11: Leveraging Pure Transformer for Multi-Class Propaganda Detection}

\author{Dmitry Grigorev \\
  Innopolis University \\
 {\tt d.grigorev@innopolis.ru} \\\And
  Vladimir Ivanov \\
  Innopolis University \\
  {\tt v.ivanov@innopolis.ru} \\}

\date{}

\begin{document}
\maketitle
\begin{abstract}
    The paper presents the solution of team "Inno" to a SEMEVAL 2020 task 11 "Detection of propaganda techniques in news articles". The goal of the second subtask is to classify textual segments that correspond to one of the 18 given propaganda techniques in news articles dataset. We tested a pure Transformer-based model with an optimized learning scheme on the ability to distinguish propaganda techniques between each other. Our model showed $0.6$ and $0.58$ overall F1 score on validation set and test set accordingly and non-zero F1 score on each class on both sets.
\end{abstract}

\section{Introduction}
\blfootnote{This work is licensed under a Creative Commons Attribution 4.0 International Licence. Licence details: http://creativecommons.org/licenses/by/4.0/.}
Modern society is experiencing an overflow of different kinds of information. Thousands of news media are publishing a giant number of articles, blogposts, twits and other kind of content every day and even every hour. People who want to stay informed about global situation are often not able to distinguish which source is reliable and provides relevant content. Therefore, such topics as automatic fake news detection, sentiment analysis and propaganda detection are gaining increasing attention.

The authors of \cite{jowett2018propaganda} defined propaganda in the most neutral sense as dissemination or promotion of a particular idea. 

This paper provides our solution to SEMEVAL 2020 task 11 ``Detection of propaganda techniques in news articles''. The task consists of two subtasks: (i) a propaganda sample span identification and bounding and (ii) a classification of these spans into one of the 18 classes (or techniques). Some techniques in the second subtask were merged into one class to get 14 classes for prediction in the end. All the techniques, as well as the other details about the task, are described by the task organizers in \cite{DaSanMartinoSemeval20task11}.

In this paper we present a solution to the second subtask. The suggested system\footnote{Source code can be found here \url{https://github.com/oxxford/SEMEVAL-2020}} consists of a Transformer-based classifier with additional learning optimization techniques such as undersampling, cost-sensitive learning and context addition, aimed at robustness. This approach allowed us to achieve $0.6$ and $0.58$ overall F1 scores with non-zero F1 score for each class on validation set and test set accordingly. We took the 7th place out of 31 teams on the second subtask. 

\section{Background}

One of the first classifications of propaganda techniques was proposed in 1936 by Clyde R. Miller \cite{edwards1938group} after a presidential election in the USA. This categorization consists of description of 7 propaganda techniques: ``Name calling'', ``Glittering generalities'', ``Transfer'', ``Testimonial'', ``Plain folks'', ``Card stacking'' and ``Band wagon''. Since then a lot more techniques have been identified, but this classification is still relevant (some of the techniques have been included in the SEMEVAL 2020).
Probably, the most promising state-of-art strategies of dealing with propaganda and filtering it from sentimentally neutral texts is the one of viewing the task as a text classification problem. Some of the research efforts in propagandist texts classification are described further in this section.         


\cite{Rashkin} investigated news sources of four categories - truthful, propagandist, hoax and satiric - to show what language features correspond to each news source category. They examined 50 highest weighted n-gram features in the MaxEntropy classifier for each class and showed that truthful sources heavily use specific places (e.g. ``London'') as well as exact time (e.g. ``on Tuesday''). For propagandist news features the sources employ abstract generalities (e.g. ``truth'', ``freedom'') and specific issues (e.g. ``election'', ``shooting''), while for satiric articles vaguely facetious hearsay (e. g. ``reportedly'', ``confirmed'') are employed. Finally, hoax publishers utilize divisive topics (e.g. ``liberals'', ``trump'') and dramatic cues (e.g. ``breaking''). The authors also proposed that hoax and propagandist sources often use videos as a reference since words like ``video'' and ``youtube'' are also included in the highest weighted for these categories.

\cite{barron2019proppy} have significantly contributed to the field. They not only collected a novel dataset that overall contains more than 35000 news articles both with and without propagandist content, but also compared what text features allow to efficiently classify articles on propagandist and non-propagandist. As a result they concluded that stylistic features like character n-grams show better results than topic-related features. The authors also demonstrated a service for news articles propaganda level assessment. 

\cite{martino2019fine} have proposed another corpus for propaganda classification. It consists of 451 articles from 48 sources. Every article was manually annotated sentence by sentence by identifying the presence or absence of one of 18 propaganda techniques (grouped into 14 classes). The authors also presented baseline models for binary and multi-class classification based on BERT \cite{devlin2018bert} and launched a task on propaganda detection that will be described later.

A major step in the field of propaganda detection was an establishment of Fine-Grained Propaganda Detection challenge \cite{EMNLP19DaSanMartino}. This challenge, similar to the current one, had two tasks:
\begin{itemize}
\item The first one was to classify sentences into propaganda and non-propaganda classes. The sentences were taken from the articles from propagandist and non-propagandist sources. The metric was F1 score.
\item The second task consisted of two parts - extract spans that contain propaganda from article text and classify what propaganda technique is used. The metric took into account both span borders and class predictions. In the current version of the challenge the task was presented by two separate tasks, span identification and classification.
\end{itemize}

Overall, 25 teams submitted their solutions of the first task and 12 teams submitted their solutions of the second task. 
Leading teams used Transformer-based \cite{vaswani2017attention} models that are built on a powerful encoder-decoder architecture having self-attention mechanism as its core. This mechanism allows to take into account inter-sequence word dependencies, providing rich semantic representation. Almost all teams used some Transformer-based models (especially BERT \cite{devlin2018bert}) either to get embeddings or as a pretrained model \cite{yoosuf-yang-2019-fine} \cite{hou-chen-2019-caunlp}. Other teams often used ensembles with different features and models inside: LSTM-CRF \cite{gupta-etal-2019-neural}, XGBoost \cite{tayyarmadabushietal2019cost}, BiLSTM \cite{vlad-etal-2019-sentence} and others. Table \ref{table1} summarizes approaches extensively used in the solutions.

\begin{table}
    \centering
    \captionsetup{justification=centering}
    \begin{tabular}{ |c|c|c| } 
    \hline
    Approach & Times it was used \\
    \hline
    BERT & 13 \\ 
    LSTM & 7 \\ 
    Language features & 5 \\
    ELMo & 3 \\ 
    Logistic regression & 3 \\
    \hline
    \end{tabular}
    \caption{How often different approaches were used}
    \label{table1}
\end{table}

As Table \ref{table1} shows, Transformer-based models (BERT in particular) have become a leading architecture in propaganda detection field.

\section{System overview}

Considering the top-performing systems of the previous challenge, we decided to use Transformer-based model as a solution baseline - mainly focusing on BERT \cite{devlin2018bert} and RoBERTa \cite{liu2019roberta}. However, since a lot of new models have been published since the previous challenge, we also tested several different Transformer-based models, including  DistilBERT \cite{sanh2019distilbert}, ALBERT \cite{lan2019albert}, XLnet \cite{yang2019xlnet}. After calculating F1 scores for all of them, we chose the one with the highest performance - RoBERTa. We also decided to use a model pretrained on a huge dataset and fine-tune it on the given dataset by adding a fully-connected classification layer. This strategy is highly suitable for our case because the training dataset consisted of 371 articles and contained 6129 propaganda samples in total for all classes, which is not enough for training a good-performing Transformer model from scratch, but is reasonable for fine-tuning. The following sections describe the techniques we used for optimizing the robustness of our model. 

\subsection{Imbalanced Data and Undersampling}

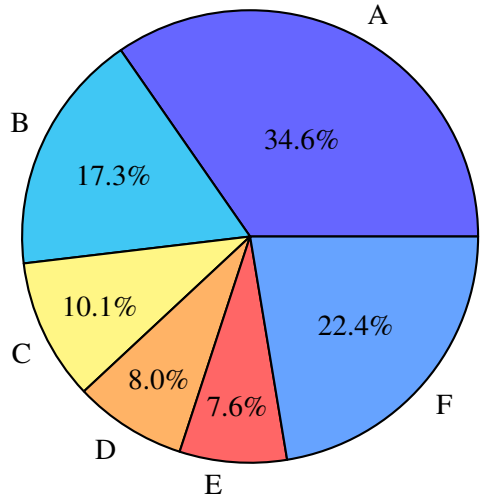
\begin{figure}
\centering
\begin{tikzpicture} 
    \pie{34.6/A, 
        17.3/B, 
        10.1/C, 
        8.0/D, 
        7.6/E, 
        22.4/F}
\end{tikzpicture}
\caption{Class distribution in the train data, where A is ``Loaded Language'', B is ``Name Calling or Labeling'', C is ``Repetition'', D is ``Doubt'', E is ``Exaggeration or Minimisation'', and F represents all the remaining 9 classes.}
\label{chart1}
\end{figure}

One essential feature of the dataset is high class imbalance. Figure \ref{chart1} demonstrates class distribution among samples.

As our first attempt to overcome class imbalance we used undersampling technique, which implies reducing the portion of more frequent class in the data by removing some samples. In our case, the most frequent classes were 'Loaded Language' (2123 out of 6129 samples in train set), and 'Name Calling,Labeling' (1058 out of 6129 samples in train set). Our aim was to retain approximately few hundred samples for each of these classes. To obtain this we experimented with different coefficients: 0.35, 0.3, 0.2 for 'Loaded Language' and 0.4, 0.5, 0.6 for 'Name Calling, Labeling'. The best combination with respect to final F1 score was 0.2 for 'Loaded Language' and 0.5 for 'Name Calling,Labeling'.

\subsection{Cost-sensitive learning}

Another technique we tried to deal with class imbalance was cost-sensitive learning. The idea of this method is to assign different "cost" to classes with different inclusion percentage in the dataset. We implemented the method by assigning label weights during loss calculation. The weight $w_i$ for each label was calculated using formula (formula \ref{formula1}):

\begin{center} 
  $w_i = \frac{c_i}{\sum\limits_{j}{c_j}}, c_i = \frac{\sum\limits_{j}{l_j}}{l_i}$
  \label{formula1}
\end{center}

where $l_i$ is number of samples of label $i$ in the dataset. This formula (formula \ref{formula1}) can be interpreted like this: class weight is inversely proportional to the portion of this class in the dataset.

\subsection{Context addition}

\begin{figure}[h] 
\centering
\captionsetup{justification=centering}
\includegraphics[width=0.85\textwidth]{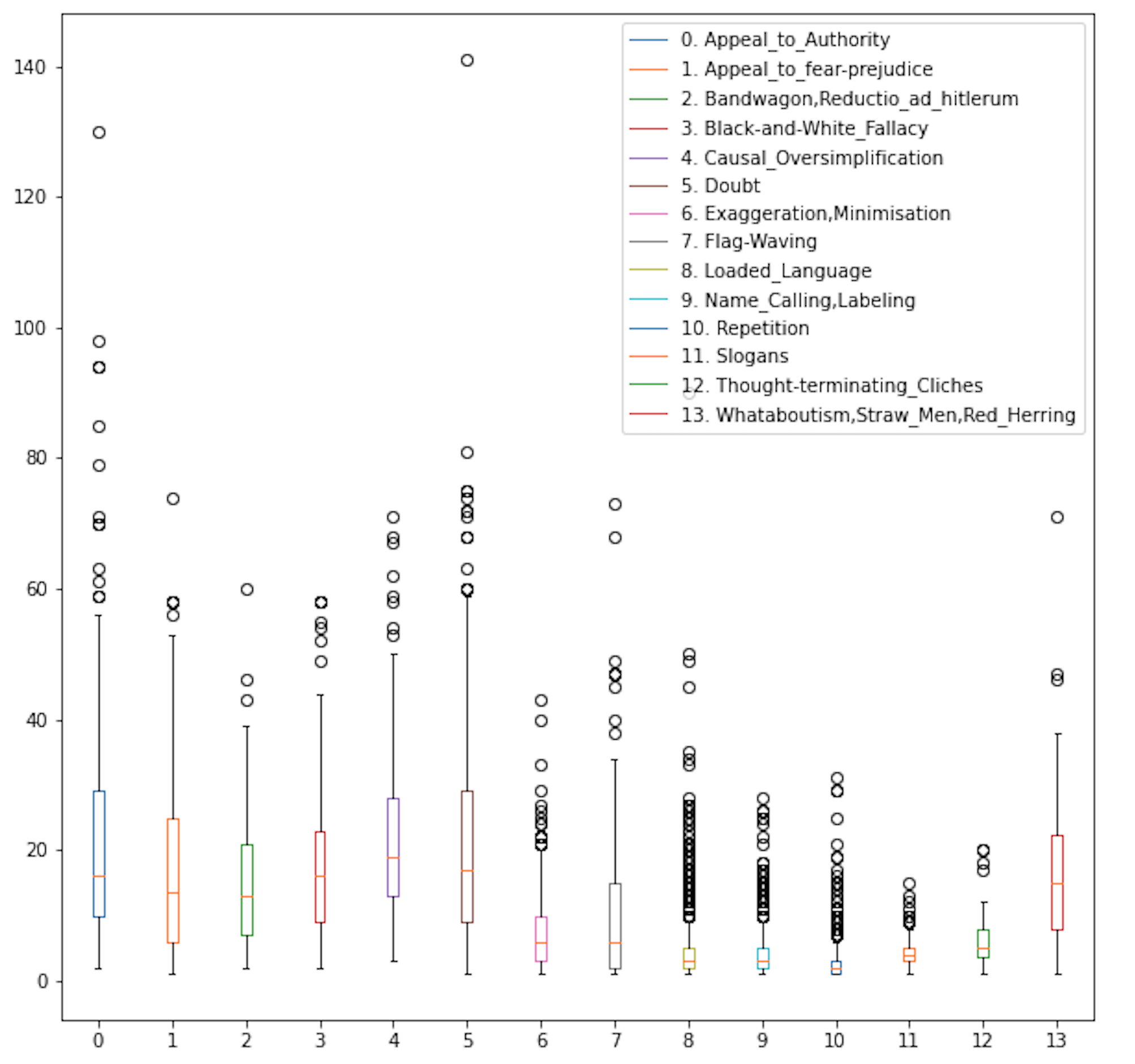}
\caption{Box-plots of length distribution in words inside classes}
\label{figure1}
\end{figure}

Having performed the dataset analysis we found out that span samples may highly vary in length: from one-word long samples to several sentences long sentences. Moreover, this significant difference may occur within the same class. Figure \ref{figure1} shows box-plots of sample lengths grouped by classes. It is interesting that three most frequent classes (8, 9, 10) have the shortest average lengths.

Our expectation was that spans with such a diverse lengths cannot efficiently represent all techniques and that some useful information can be left outside the span boundaries. To check this, we decided to expand the spans with the surrounding context. We tried two ways of setting the new sample boundaries: to complete a sentence (any text surrounded by \textit{'.', '?', '!'} signs or "end of line" symbol on both sides) and to complete a sentence or a subsentence (same as complete a sentence but with \textit{','} sign included). However, this spans expansion strategy was not efficient as it introduced more noise and, thus, reduced the score.

\section{Experimental Setup and Results}

To choose between different Transformer-based models we trained each one of them on a single epoch and compared their validation scores.  The result of this evaluation are shown in Table \ref{table4}.

\begin{table}
    \centering
    \captionsetup{justification=centering}
    \begin{tabular}{ |c|c|c| } 
    \hline
    Model & F1 val. score  \\
    \hline
    ALBERT & 0.498 \\ 
    DistillBERT & 0.502 \\ 
    XLnet & 0.526 \\ 
    BERT & 0.553 \\ 
    \textbf{RoBERTa} & \textbf{0.573} \\ 
    \hline
    \end{tabular}
    \caption{Baseline models performance}
    \label{table4}
\end{table}

According to Table \ref{table4}, RoBERTa showed the best performance results among all the other models, therefore we chose it as a model for future improvement.
It is also worth mentioning that all models showed similar patterns on the same classes:

\begin{itemize}
    \item High performance ($F_1 \geq  0.5$) on "Flag-Waving", "Loaded Language" and "Name Calling, Labeling";
    \item Intermediate performance ($0.1 \leq F_1 < 0.5$) on "Appeal to fear-prejudice", "Doubt", "Exaggeration, Minimisation" and "Repetition";
    \item Low performance ($F_1 < 0.1$) on "Appeal to Authority", "Bandwagon, Reductio ad hitlerum", "Black-and-White Fallacy", "Causal Oversimplification", "Slogans", "Thought-terminating Cliches" and "Whataboutism, Straw Men, Red Herring".
\end{itemize}

As the next step, we tuned the hyperparameters of the model including the number of epochs, batch size and maximum sequence length. This step allowed us to improve F1 score for 2 points and provided us with a ready-to-run model so that we could proceed to optimization techniques described in previous section. Table \ref{table2} shows the result of applying this strategy.
 
\begin{table} 
    \centering
    \captionsetup{justification=centering}
    \begin{tabular}{ |c|c|c| } 
    \hline
    Strategy & F1 val. score  \\
    \hline
    Baseline & 0.57384 \\ 
    Tuned & 0.59172 \\ 
    Tuned+Undersampling & 0.59266 \\ 
    \textbf{Tuned+Cost-sensitive learning} & \textbf{0.60113} \\ 
    Tuned+Context addition & 0.55691 \\ 
    \hline
    \end{tabular}
    \caption{Performance with different techniques applied}
    \label{table2}
\end{table}

\subsection{Our submission}

Our final system submitted for this task is based on RoBERTa model, combined with cost-sensitive learning strategy. We took pretrained weights from Transformers \cite{Wolf2019HuggingFacesTS} library by Hugging Face, and fine-tuned the model on 5 epochs with batch size 8 and maximum sequence length of 512. We used Simple Transformers\footnote{\url{https://github.com/ThilinaRajapakse/simpletransformers}} library for training and inference.

The submitted model showed $0.58$ F1 score on the test set taking the 7th place among 31 teams in the second track. The system also showed non-zero F1 score on each class (some models with better final performance did not achieve this), which means that the model can capture some meaning of each propaganda technique. It is also worth mentioning that the difference between validation performance and test performance of our model is 2 points, which is less than that of most models with better test performance and, moreover, it is less than that of models in top three. Thus, we achieved less overfit compared to the competitors.

\section{Conclusion}
In this paper we described a system that we developed to participate in SEMEVAL 2020 Task 11 ``Detection of propaganda techniques in news articles''. We trained a pure Transformer-based model with preprocessing and achieved $0.58$ F1 score on test data and non-zero F1 score for each class. The approach demonstrated the effectiveness of combining a Transformer-based baseline with learning optimization techniques including undersampling, cost-sensitive learning and context augmentation.

We also demonstrated that compared to other NLP tasks that are being actively solved, propaganda techniques classification task is far behind, and both models and data are to be improved. However, non-zero scores for each class show that various baseline models are able to capture propagandist semantics and stronger models for this field can be built.

\bibliographystyle{coling}
\bibliography{coling2020}

\begin{thebibliography}{}

\bibitem[\protect\citename{Barr{\'o}n-Cedeno \bgroup et al.\egroup
  }2019]{barron2019proppy}
Alberto Barr{\'o}n-Cedeno, Israa Jaradat, Giovanni Da~San~Martino, and Preslav
  Nakov.
\newblock 2019.
\newblock Proppy: Organizing the news based on their propagandistic content.
\newblock {\em Information Processing \& Management}.

\bibitem[\protect\citename{Da~San~Martino \bgroup et al.\egroup
  }2019]{EMNLP19DaSanMartino}
Giovanni Da~San~Martino, Seunghak Yu, Alberto Barr\'{o}n-Cede\~no, Rostislav
  Petrov, and Preslav Nakov.
\newblock 2019.
\newblock Fine-grained analysis of propaganda in news articles.
\newblock In {\em Proceedings of the 2019 Conference on Empirical Methods in
  Natural Language Processing and the 9th International Joint Conference on
  Natural Language Processing, EMNLP-IJCNLP 2019}, EMNLP-IJCNLP 2019, Hong
  Kong, China, November.

\bibitem[\protect\citename{Da~San~Martino \bgroup et al.\egroup
  }2020]{DaSanMartinoSemeval20task11}
Giovanni Da~San~Martino, Alberto Barr\'{o}n-Cede\~no, Henning Wachsmuth,
  Rostislav Petrov, and Preslav Nakov.
\newblock 2020.
\newblock {SemEval}-2020 task 11: Detection of propaganda techniques in news
  articles.
\newblock In {\em Proceedings of the 14th International Workshop on Semantic
  Evaluation}, SemEval 2020, Barcelona, Spain, September.

\bibitem[\protect\citename{Devlin \bgroup et al.\egroup }2018]{devlin2018bert}
Jacob Devlin, Ming-Wei Chang, Kenton Lee, and Kristina Toutanova.
\newblock 2018.
\newblock Bert: Pre-training of deep bidirectional transformers for language
  understanding.
\newblock {\em arXiv preprint arXiv:1810.04805}.

\bibitem[\protect\citename{Edwards}1938]{edwards1938group}
Violet Edwards.
\newblock 1938.
\newblock {\em Group leader's guide to propaganda analysis: Revised edition of
  experimental study materials for use in junior and senior high schools, in
  college and university classes, and in adult study groups}.
\newblock Institute for propaganda analysis, Incorporated.

\bibitem[\protect\citename{Gupta \bgroup et al.\egroup
  }2019]{gupta-etal-2019-neural}
Pankaj Gupta, Khushbu Saxena, Usama Yaseen, Thomas Runkler, and Hinrich
  Sch{\"u}tze.
\newblock 2019.
\newblock Neural architectures for fine-grained propaganda detection in news.
\newblock In {\em Proceedings of the Second Workshop on Natural Language
  Processing for Internet Freedom: Censorship, Disinformation, and Propaganda},
  pages 92--97, Hong Kong, China, November. Association for Computational
  Linguistics.

\bibitem[\protect\citename{Hou and Chen}2019]{hou-chen-2019-caunlp}
Wenjun Hou and Ying Chen.
\newblock 2019.
\newblock {CAU}n{LP} at {NLP}4{IF} 2019 shared task: Context-dependent {BERT}
  for sentence-level propaganda detection.
\newblock In {\em Proceedings of the Second Workshop on Natural Language
  Processing for Internet Freedom: Censorship, Disinformation, and Propaganda},
  pages 83--86, Hong Kong, China, November. Association for Computational
  Linguistics.

\bibitem[\protect\citename{Jowett and O'donnell}2018]{jowett2018propaganda}
Garth~S Jowett and Victoria O'donnell.
\newblock 2018.
\newblock {\em Propaganda \& persuasion}.
\newblock Sage Publications.

\bibitem[\protect\citename{Lan \bgroup et al.\egroup }2019]{lan2019albert}
Zhenzhong Lan, Mingda Chen, Sebastian Goodman, Kevin Gimpel, Piyush Sharma, and
  Radu Soricut.
\newblock 2019.
\newblock Albert: A lite bert for self-supervised learning of language
  representations.
\newblock {\em arXiv preprint arXiv:1909.11942}.

\bibitem[\protect\citename{Liu \bgroup et al.\egroup }2019]{liu2019roberta}
Yinhan Liu, Myle Ott, Naman Goyal, Jingfei Du, Mandar Joshi, Danqi Chen, Omer
  Levy, Mike Lewis, Luke Zettlemoyer, and Veselin Stoyanov.
\newblock 2019.
\newblock Roberta: A robustly optimized bert pretraining approach.
\newblock {\em arXiv preprint arXiv:1907.11692}.

\bibitem[\protect\citename{Martino \bgroup et al.\egroup
  }2019]{martino2019fine}
Giovanni Da~San Martino, Seunghak Yu, Alberto Barr{\'o}n-Cedeno, Rostislav
  Petrov, and Preslav Nakov.
\newblock 2019.
\newblock Fine-grained analysis of propaganda in news articles.
\newblock {\em arXiv preprint arXiv:1910.02517}.

\bibitem[\protect\citename{Rashkin \bgroup et al.\egroup }2017]{Rashkin}
Hannah Rashkin, Eunsol Choi, Jin Jang, Svitlana Volkova, and Yejin Choi.
\newblock 2017.
\newblock Truth of varying shades: Analyzing language in fake news and
  political fact-checking.
\newblock pages 2931--2937, 01.

\bibitem[\protect\citename{Sanh \bgroup et al.\egroup
  }2019]{sanh2019distilbert}
Victor Sanh, Lysandre Debut, Julien Chaumond, and Thomas Wolf.
\newblock 2019.
\newblock Distilbert, a distilled version of bert: smaller, faster, cheaper and
  lighter.
\newblock {\em arXiv preprint arXiv:1910.01108}.

\bibitem[\protect\citename{Tayyar~Madabushi \bgroup et al.\egroup
  }2019]{tayyarmadabushietal2019cost}
Harish Tayyar~Madabushi, Elena Kochkina, and Michael Castelle.
\newblock 2019.
\newblock Cost-sensitive {BERT} for generalisable sentence classification on
  imbalanced data.
\newblock In {\em Proceedings of the Second Workshop on Natural Language
  Processing for Internet Freedom: Censorship, Disinformation, and Propaganda},
  pages 125--134, Hong Kong, China, November. Association for Computational
  Linguistics.

\bibitem[\protect\citename{Vaswani \bgroup et al.\egroup
  }2017]{vaswani2017attention}
Ashish Vaswani, Noam Shazeer, Niki Parmar, Jakob Uszkoreit, Llion Jones,
  Aidan~N Gomez, {\L}ukasz Kaiser, and Illia Polosukhin.
\newblock 2017.
\newblock Attention is all you need.
\newblock In {\em Advances in neural information processing systems}, pages
  5998--6008.

\bibitem[\protect\citename{Vlad \bgroup et al.\egroup
  }2019]{vlad-etal-2019-sentence}
George-Alexandru Vlad, Mircea-Adrian Tanase, Cristian Onose, and
  Dumitru-Clementin Cercel.
\newblock 2019.
\newblock Sentence-level propaganda detection in news articles with transfer
  learning and {BERT}-{B}i{LSTM}-capsule model.
\newblock In {\em Proceedings of the Second Workshop on Natural Language
  Processing for Internet Freedom: Censorship, Disinformation, and Propaganda},
  pages 148--154, Hong Kong, China, November. Association for Computational
  Linguistics.

\bibitem[\protect\citename{Wolf \bgroup et al.\egroup
  }2019]{Wolf2019HuggingFacesTS}
Thomas Wolf, Lysandre Debut, Victor Sanh, Julien Chaumond, Clement Delangue,
  Anthony Moi, Pierric Cistac, Tim Rault, R'emi Louf, Morgan Funtowicz, and
  Jamie Brew.
\newblock 2019.
\newblock Huggingface's transformers: State-of-the-art natural language
  processing.
\newblock {\em ArXiv}, abs/1910.03771.

\bibitem[\protect\citename{Yang \bgroup et al.\egroup }2019]{yang2019xlnet}
Zhilin Yang, Zihang Dai, Yiming Yang, Jaime Carbonell, Ruslan Salakhutdinov,
  and Quoc~V Le.
\newblock 2019.
\newblock Xlnet: Generalized autoregressive pretraining for language
  understanding.
\newblock {\em arXiv preprint arXiv:1906.08237}.

\bibitem[\protect\citename{Yoosuf and Yang}2019]{yoosuf-yang-2019-fine}
Shehel Yoosuf and Yin Yang.
\newblock 2019.
\newblock Fine-grained propaganda detection with fine-tuned {BERT}.
\newblock In {\em Proceedings of the Second Workshop on Natural Language
  Processing for Internet Freedom: Censorship, Disinformation, and Propaganda},
  pages 87--91, Hong Kong, China, November. Association for Computational
  Linguistics.

\end{thebibliography}

\end{document}